\documentclass[hidelinks,hypertexnames=false,sn-mathphys,Numbered,oneside]{sn-jnl}

\usepackage{newtxtext}
\usepackage{manyfoot}
\usepackage{xurl}           
\usepackage{graphicx}%
\usepackage{multirow}%
\usepackage{amsmath,amssymb,amsfonts}%
\usepackage{newtxmath}
\usepackage{mathrsfs}%
\usepackage[title]{appendix}%
\usepackage{xcolor}%
\usepackage{textcomp}%
\usepackage{booktabs}%
\usepackage{algorithm}%
\usepackage{algorithmicx}%
\usepackage{algpseudocode}%
\usepackage{listings}%
\usepackage{subcaption}
\usepackage{array,booktabs}
\usepackage{siunitx}
\usepackage{tabularx}
\newcolumntype{L}{>{\raggedright\arraybackslash}X}   
\begin{document}

\title{DamageCAT:\ A Deep Learning Transformer Framework for Typology-Based Post-Disaster Building Damage Categorization}


\author*[1]{\fnm{Yiming} \sur{Xiao}}\email{yxiao@tamu.edu}

\author[1]{\fnm{Ali} \sur{Mostafavi}}\email{amostafavi@civil.tamu.edu}

\affil*[1]{\orgdiv{Zachry Department of Civil and Environmental Engineering}, \orgname{Texas A\&M University}, \orgaddress{\street{199 Spence St}, \city{College Station}, \postcode{TX 77840}, \state{Texas}, \country{USA}}}

\abstract{Rapid, accurate, and descriptive building damage assessment is critical for directing post-disaster resources, yet current automated methods typically provide only binary (damaged/undamaged) or ordinal severity scales. This paper introduces DamageCAT, a framework that advances damage assessment through typology-based categorical classifications. We contribute: (1) the BD-TypoSAT dataset containing satellite image triplets from Hurricane Ida with four damage categories—partial roof damage, total roof damage, partial structural collapse, and total structural collapse—and (2) a hierarchical U-Net-based transformer architecture for processing pre- and post-disaster image pairs. Our model achieves 0.737 IoU and 0.846 F1-score overall, with cross-event evaluation demonstrating transferability across Hurricane Harvey, Florence, and Michael data. While performance varies across damage categories due to class imbalance, the framework shows that typology-based classification can provide more actionable damage assessments than traditional severity-based approaches, enabling targeted emergency response and resource allocation.}

\keywords{Damage assessment, Satellite imagery, Transformers, Damage description}

\maketitle

\section{Introduction}\label{sec:intro}
    \subsection{Motivation \& problem statement}
    The increasing frequency and severity of natural disasters pose significant challenges to emergency response systems worldwide. According to \citet{gholami_deployment_2022}, natural disasters affect approximately 350 million people annually, causing billions of dollars in damages. Climate change is expected to further exacerbate these trends, leading to more frequent and intense disasters \citep{PIZZORNI2024100169}. In this context, rapid and accurate building damage assessment becomes crucial for effective disaster response and recovery.
    Building damage assessment serves multiple critical purposes in disaster response. As \citet{adriano_developing_2023} emphasizes, timely damage assessment is essential for both emergency response and prompt recovery efforts. Similarly, \citet{dias_conditional_2024} highlights that building damage assessment is vital for guiding disaster response missions and estimating recovery needs across impacted areas. The information gathered through damage assessment enables humanitarian organizations to be prompt, effective, and efficient in supporting affected populations \citep{bouchard_transfer_2022}.
    Traditional damage assessment methods rely heavily on manual visual interpretation of post-disaster imagery. However, this approach presents significant limitations.\ \citet{alisjahbana_deepdamagenet_2024} point out that manual interpretation is not only time-consuming but often results in very low accuracy. The labor-intensive and tedious nature of manual annotation further hampers rapid response efforts \citep{bouchard_transfer_2022}. In addition, field-based assessments may be dangerous or impossible in severely affected areas, creating information gaps in critical locations.
    Satellite imagery offers a promising alternative for building damage assessment. In recent years, the use of remote sensing data and machine learning has significantly enhanced damage assessment in disasters.\ \citet{kahl_towards_2024} note that satellites provide a unique, comprehensive viewpoint for information retrieval from disaster-affected zones. Commercial satellite providers now enable near-daily monitoring of Earth's surface, facilitating timely post-disaster assessments \citep{dias_conditional_2024}. Moreover, remote sensing technologies provide rich and reliable information to support expert decision-making \citep{bouchard_transfer_2022}.
    Despite these advantages, satellite-based damage assessment faces several challenges.\ \citet{chen_hrtbda_2024} observed that the task is complicated by diverse building structures and complex environments.\ \citet{liu_novel_2022} highlight that building damage classification is challenging due to limited publicly available imagery data of damaged buildings and the difficulty in judging damage levels from images. Furthermore, most existing deep learning models are designed for non-damaged building segmentation and may not be suitable for damaged building classification \citep{liu_novel_2022}.
    The integration of deep learning with satellite imagery has shown promising results in automating building damage assessment.\ \citet{neto_building_2024} demonstrated the potential of combining mathematical morphology and convolutional neural networks for building damage segmentation. Similarly, \citet{may_building_2022} utilized advanced networks such as EfficientNet and Siamese models to detect buildings and evaluate damage levels using pre- and post-event images.

    However, several limitations persist in current deep-learning approaches for remote sensing-based disaster-damaged assessment.\ \citet{melamed_uncovering_2024} identified bias in commonly used datasets, noting that deep learning models struggle to accurately identify isolated damaged buildings, potentially causing oversights in critical disaster scenarios and delaying humanitarian aid. In addition, \citet{wang_building_2022} point out challenges related to extremely imbalanced datasets, where damaged buildings are significantly underrepresented compared to undamaged ones. Another significant limitation of current deep learning-based models for building damage assessment is their reliance on subjective severity scales (e.g., ``low,'' ``moderate,'' ``high''), which offer limited insight into the specific nature and type of damage. These severity labels often arise from coarse-grained assessments that can vary between datasets or annotators, leading to inconsistencies and ambiguity. Moreover, a single severity score cannot distinguish whether damage is confined to a roof or extends into load-bearing structures, nor can it differentiate between partial and total collapse. As a result, key nuances---such as the potential for secondary hazards or the requisite expertise for repairs---are frequently overlooked. This lack of descriptive granularity ultimately hampers targeted resource allocation, informed decision-making, and effective prioritization of rescue and recovery efforts in post-disaster scenarios.
    
    Despite the growing amount of literature leveraging deep learning for post-disaster damage assessments, current severity-based classification approaches often provide only coarse estimates. While previous studies have made important strides in damage classification, subjective rating scales can neglect the underlying mechanisms or precise nature of the damage, potentially limiting model applicability for decision-making in real disaster scenarios. Knowing whether a structure exhibits partial roof damage or total structural collapse is not just a matter of granularity; it directly influences emergency response strategies, resource allocation, and subsequent recovery efforts. For instance, responding to a partially damaged roof may call for specialized roofing repair teams and temporary protective measures, whereas a structural collapse demands immediate search-and-rescue operations and comprehensive rebuilding plans. Building upon foundational work in remote sensing-based damage assessment, this study explores a typology-based classification approach that aims to capture distinct modes of structural failure, complementing existing automated assessment systems. This research context emphasizes opportunities to enhance damage evaluation across various disaster conditions, construction types, and geographic regions, while acknowledging the inherent challenges of satellite-based assessment.

    \subsection{Research objectives}   
        This study aims to contribute to building damage assessment by exploring typology-based classification framework that captures distinct modes of structural failure. We present a deep learning architecture that processes pre- and post-disaster satellite imagery to identify and categorize building damage with greater detail. To support this exploration, we introduce BD-TypoSAT (Building Damage Typology Satellite Dataset), containing four specific damage categories---partial roof damage, total roof damage, partial structural collapse, and total structural collapse---offering potentially richer representation than standard severity-based labels. We then deploy a transformer-driven, hierarchical U-Net that integrates semantic segmentation and change detection to segment building footprints and classify them by damage type. This approach aims to yield actionable information for emergency managers and stakeholders. Finally, we rigorously evaluate the model's performance using multiple metrics (e.g., IoU, F1) to demonstrate its utility under real-world conditions, and we discuss both the operational implications and limitations for disaster response, such as more informed resource allocation and prioritization of recovery efforts.

    \subsection{Contributions}
        This study makes several key contributions to automated building damage assessment. We introduce the BD-TypoSAT dataset, derived from high-resolution satellite imagery captured after Hurricane Ida (2021), which categorizes building damage into four distinct typological classes—partial roof damage, total roof damage, partial structural collapse, and total structural collapse. This typological approach offers a more nuanced alternative to traditional binary or coarse severity classifications, enabling targeted decision support by distinguishing between surface-level roof damage and deeper structural failures.

        Our computational framework integrates a hierarchical U-Net-based transformer architecture with multi-scale difference blocks that jointly process pre- and post-disaster satellite images. The transformer backbone facilitates long-range contextual learning while preserving critical spatial detail through the U-Net structure. At each resolution level, difference blocks explicitly compare corresponding features between temporal images, enabling detection of subtle structural changes. To address the severe class imbalance inherent in disaster data, we implement a weighted combination of cross-entropy and Dice losses, ensuring that rarer damage categories such as total structural collapse are not overlooked during training.

        Quantitative validation demonstrates promising performance with an overall IoU of 0.737 and F1-score of 0.846, while cross-event evaluation on Hurricanes Harvey, Florence, and Michael shows encouraging transferability across different disaster contexts. Though accuracy varies across urban environments and damage types, the framework provides actionable insights for emergency response applications. By releasing both the BD-TypoSAT dataset and our transformer-driven architecture, this research contributes valuable resources for future studies aiming to expand typological damage classification to other hazards or incorporate additional data modalities.

    \subsection{Paper organization}
        First in Section~\ref{sec:2}, we review some previous work that is highly related to this paper. Later we introduce the method, including the dataset, experimental setup, and the model training details, in Section~\ref{sec:3}. In Section~\ref{sec:5}, we discuss the proposed framework's significance and list limitations and challenges. Finally, we conclude the work and give future research suggestions in Section~\ref{sec:6}.

\section{Related work}\label{sec:2}
Research on automated building damage assessment has rapidly evolved, leveraging a wide variety of remote sensing, GIS-based techniques, and deep learning.

In the pre-deep learning era, remote sensing imagery was recognized as a valuable tool for rapid post-disaster damage mapping, allowing for swift assessment over wide areas \citep{kerle_satellite-based_2010}, offering advantages like efficiency and cost-effectiveness compared to field surveys. For detailed structural damage assessment at the individual building level, \citet{womble_remote_2008} highlighted the necessity of very high spatial resolution imagery, typically sub-meter pixel size, enabling the generation of various map products focusing on individual buildings, city blocks, or aggregated areas, often through visual interpretation by experienced analysts. However, this foundational approach faced significant limitations.\ \citet{kerle_satellite-based_2010} noted that imagery quality can be poor or insufficient, and clouds frequently obscure affected areas. The predominantly vertical perspective of satellite imagery, as pointed out by \citet{yamazaki_visual_2005}, makes damage visible primarily on roofs, often missing facade damage or interior issues, especially less severe ones, and making it difficult to distinguish between different damage grades. This ambiguity was further compounded by a lack of universally accepted nomenclature for image-mapped damage \citep{kerle_satellite-based_2010}. Moreover, accurately assessing map quality proved challenging due to a scarcity of suitable ground truth data; while field surveys offer higher accuracy, they can be incomplete or inconsistent, particularly when conducted rapidly \citep{kerle_satellite-based_2010}, and obtaining ground truth at a scale comparable to high-resolution imagery interpretation remains difficult \citep{saito_visual_2005}. These early challenges emphasized the need for more robust methods and data sources.

To address some of these limitations, integrating different data sources was explored to improve damage mapping. The use of GIS became crucial for integrating remote sensing data with other information, such as pre-disaster building databases \citep{van2000remote, womble_remote_2008}, and combining image-based mapping with limited ground validation was considered an ideal approach, particularly for areas difficult to interpret from imagery alone.\ \citet{kerle_satellite-based_2010} emphasized that using multi-temporal imagery (pre- and post-event) is often more effective for change detection related to damage than relying on post-event images alone, leading to the exploration of automated change detection techniques like threshold segmentation and change vector analysis. The development of multimodal datasets, for instance combining optical and SAR imagery, also began to support training more advanced algorithms \citep{kerle_satellite-based_2010}. However, even with these advancements, challenges related to data perspective persisted. As noted by \citet{womble_remote_2008} and \citet{yamazaki_visual_2005}, damage assessment primarily relied on nadir (vertical) perspectives, making facade damage difficult to discern. While airborne platforms offered potential for oblique views, a detailed review of analysis techniques for oblique satellite or aerial imagery remained limited in early literature \citep{womble_remote_2008}, representing a gap in comprehensive damage visibility.

With the advent of deep learning, researchers increasingly utilized change detection and temporal fusion techniques to capture pre- and post-disaster variations. For instance, \citet{zheng_building_2021} presented a deep object-based semantic change detection framework to accelerate disaster response by directly comparing pre- and post-event images. To address labeling costs, \citet{ismail_bldnet_2022} proposed BLDNet, a semi-supervised framework using graph convolutional networks and urban domain knowledge, though its reliance on graph-based relationships may falter in areas with atypical urban layouts.\ \citet{weber_building_2020} focused on fusing multi-temporal satellite imagery to enhance damage assessment in dynamic environments, but inconsistent image quality could sometimes degrade accuracy. Earlier, \citet{zhan_change_2017} employed a Siamese convolutional network to match pairwise images, boosting discriminative power, yet this approach could suffer when spatial alignment faltered. More recently, \citet{chen_changemamba_2024} introduced ChangeMamba, formulating change detection via a spatiotemporal state space model to manage uncertainties in temporal sequences; however, tuning its parameters on large datasets remains challenging. In an effort to improve generalizability, \citet{ahn_generalizable_2025} combined a vision foundation model with change detection, although it was found to struggle with highly specialized local scenarios.\ \citet{mohammadian_siamixformer_2023} proposed Siamixformer, a fully transformer-based Siamese network that fuses temporal features for fine-grained extraction, but its high memory consumption can limit real-time application. These approaches demonstrate significant progress, yet often come with trade-offs in complexity, generalizability, or computational cost.

Beyond temporal fusion, many studies have focused on enhancing network architectures to better address damage assessment tasks.\ \citet{gu_advances_2024} surveyed rapid building damage identification methods, emphasizing the efficiency of CNNs and transformer-based approaches, while critically noting that a lack of standardized datasets leads to inconsistent performance comparisons. Specific architectural innovations include the work of \citet{oludare_attention-based_2022}, who proposed an attention-based two-stream high-resolution network for satellite images; their method fuses high-resolution spatial features with attention modules for strong accuracy, but its heavy reliance on large labeled datasets limits adaptability in sparsely annotated regions.\ \citet{shen_bdanet_2022} introduced BDANet, a multi-scale CNN with cross-directional attention to capture local and global damage cues, though its high computational overhead poses challenges for resource-constrained settings. An ensemble machine learning approach by \citet{roohi_developing_2024} combined multiple classifiers to improve flood damage predictions, yet this method requires careful tuning of numerous hyperparameters.\ \citet{chen_dual-tasks_2022} developed a dual-task Siamese transformer framework for simultaneous segmentation and damage classification, which, while yielding richer features through joint optimization, also increases computational demands. Similarly, \citet{kaur_largescale_2023} designed a hierarchical transformer for large-scale assessment by organizing features at multiple scales, but balancing depth and inference speed remains difficult.\ \citet{gupta_rescuenet_2020} introduced RescueNet to unify building segmentation and damage assessment, streamlining training, though the model sometimes confused segmentation with damage levels. Foundational architectural modules such as Squeeze-and-Excitation blocks \citep{hu_squeeze-and-excitation_2018}, U-Net \citep{navab_u-net_2015}, and UNet++ \citep{zhou_unet_2018} have been widely incorporated, boosting feature recalibration and segmentation; however, these innovations often add parameters and memory requirements, complicating deployment in resource-limited settings.

Recognizing the critical challenge of labeled data scarcity, researchers have also innovated in training approaches and dataset creation.\ \citet{miglani_deep_2019} reviewed deep learning models for hazard-damaged building detection, synthesizing various strategies and highlighting the difficulty in transferring models across different disaster types. To mitigate data limitations, \citet{singh_post_2023} developed a semi-supervised transformer method leveraging ultra-high-resolution aerial imagery and unlabeled data, though it faces heavy computational loads with large-scale images.\ \citet{ro_scalable_2024} proposed a scalable method for creating annotated disaster image databases, which, while supporting model development, may underrepresent rare disaster scenarios. A self-supervised contrastive learning approach pioneered by \citet{xia_self-supervised_2022} reduces reliance on ground-truth labels, achieving performance comparable to supervised methods, but it can struggle with fine-grained damage distinctions. The release of datasets like xBD by \citet{gupta_xbd_2019}, one of the largest repositories with over 850,000 building annotations, has been crucial for benchmarking models, despite its inherent class imbalance and image quality variability. These efforts aim to make deep learning more applicable in data-scarce disaster contexts, though challenges in model transferability and dataset representativeness persist.

Further efforts to enhance damage assessment involve integrating diverse data sources for a more comprehensive picture of disaster impacts. For example, \citet{braik_automated_2024} proposed a framework integrating satellite imagery, GIS layers, and deep learning for large-scale mapping, improving building footprint extraction and damage classification; however, this approach faces challenges in real-time deployment due to multi-source data integration complexity.\ \citet{dong_integrated_2021} analyzed integrated infrastructure plans to enhance post-hazard access to critical facilities, bridging urban planning data with damage models, though this method depends on extensive data availability.\ \citet{techapinyawat_integrated_2024} applied deep learning with post-classification correction for urban land cover analysis, which can refine damage mapping but may sometimes miss atypical disaster-induced changes. Focusing on practical deployment, \citet{gholami_deployment_2022} addressed issues like limited computing resources and data sharing constraints for satellite imagery-based tools.\ \citet{xue_post-hurricane_2024} introduced a multi-modal approach fusing street-view imagery with structured building data for refined damage predictions, though its effectiveness is confined to regions with both reliable ground-level visuals and comprehensive metadata. While these integrated approaches offer richer insights, they often introduce complexities in data fusion, real-time processing, and data availability.

Despite the substantial progress reviewed, a significant gap persists in satellite-based building damage assessment: most existing methods still rely on broad severity scales, such as binary ``damaged/undamaged'' labels or imprecise ``low/medium/high'' ratings. As highlighted throughout the literature, these approaches often fail to capture the complexity of real-world damage scenarios. By focusing predominantly on severity, they can overlook critical descriptive details, such as whether damage is localized to the roof or extends into load-bearing structures—information pivotal for repair priority, resource allocation, and insurance decisions. Nuanced indicators like partial collapses or roof breaches, which can signal specific failure mechanisms and potential secondary hazards, often remain unaddressed in standard severity-centric frameworks. Consequently, decision-makers, from emergency responders to policy planners, may lack the granular insights needed to tailor interventions and expedite recovery. Addressing this shortfall necessitates a shift towards frameworks that move beyond merely measuring ``how much'' damage occurred to clarifying ``what kind'' of damage has taken place, enabling more targeted, context-aware strategies for disaster response, reconstruction, and long-term resilience planning. This study aims to contribute to filling this gap by exploring a typology-based classification approach.

\section{Method}\label{sec:3}
\begin{figure}[!htbp]
    \centering
    \includegraphics[width=\linewidth]{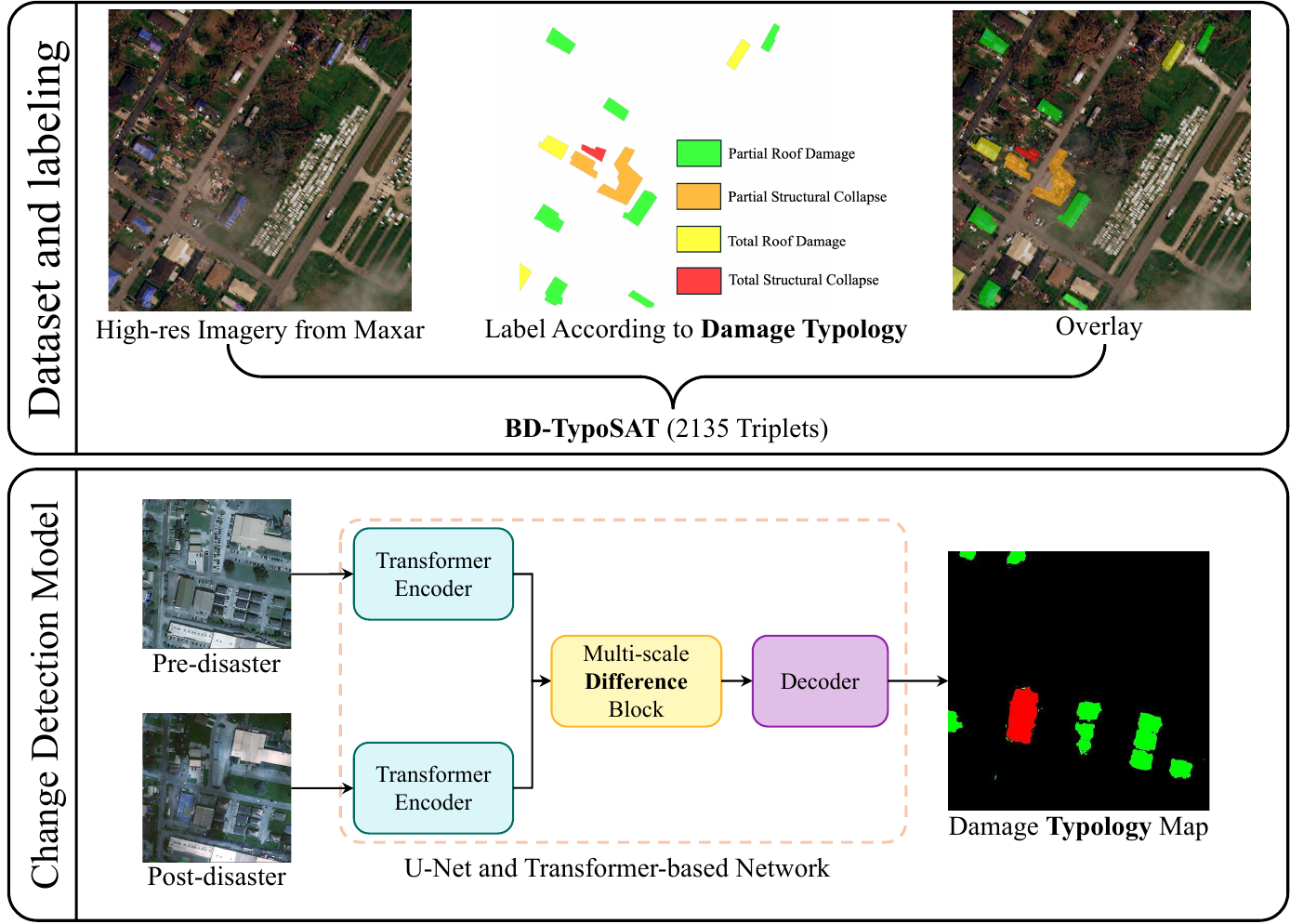}
    \caption{The overall workflow of the present study.}\label{fig:conceptual}
\end{figure}
    In this study, we build upon the hierarchical transformer architecture proposed by \citet{kaur_largescale_2023} for image change detection, tailoring it to the specific requirements of post-disaster damage assessment. The model simultaneously processes pre- and post-disaster satellite images to produce a change map that highlights structural alterations between the two temporal states. Central to this approach is the difference block, a novel component designed to extract and compare multi-scale features across each stage of the network, thereby capturing subtle yet critical variations indicative of diverse damage types. This architecture effectively combines the long-range contextual learning strengths of transformers with the detailed localization capabilities of a U-Net-like design, making it particularly suited for high-resolution satellite imagery. The overall workflow is shown in Figure~\ref{fig:conceptual}.

\subsection{Dataset}
    \begin{figure}[!htbp]
    \centering
    \includegraphics[width=0.95\linewidth]{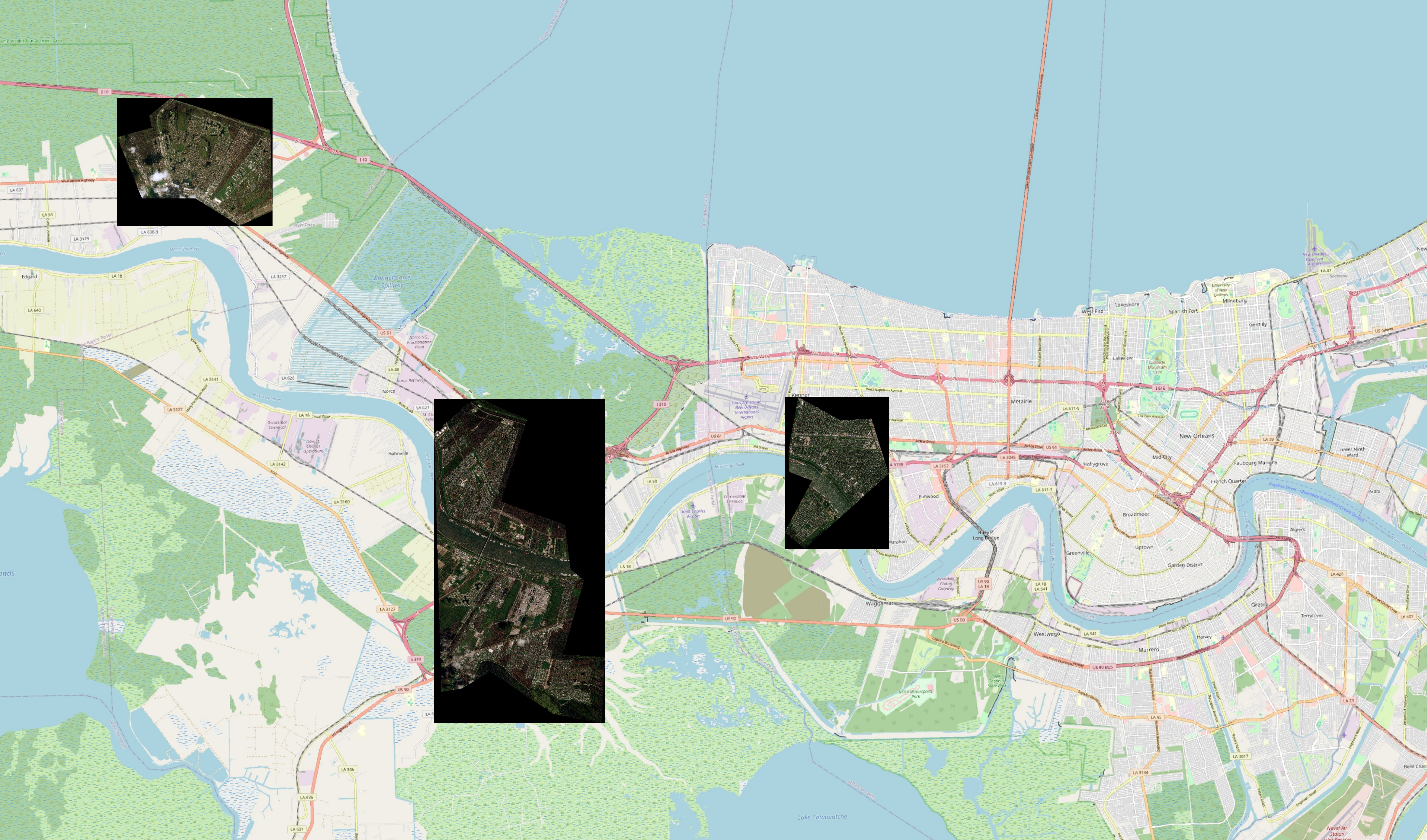}
    \caption{The study area where the satellite imagery is captured.}\label{fig:studyarea}
    \end{figure}
    
    We present the novel dataset employed for training and validating the model in this research---the BD-TypoSAT (Building Damage Typology Satellite Dataset) dataset---which was developed by us utilizing Maxar's satellite images captured during Hurricane Ida.
    On Sunday, August 29, 2021, Hurricane Ida struck parts of Louisiana and Mississippi with wind gusts reaching up to 172 mph, leaving more than a million customers without electricity, including the entire New Orleans area \citep{MaxarHurricaneIda}. During the disaster, Maxar captured high spatial resolution satellite imagery (at 0.4 m/pixel) and it was subsequently made publicly available \citep{MaxarAWSOpenData}. Figure~\ref{fig:studyarea} illustrates the region where the satellite images were obtained. The original images were segmented into 512 \(\times\) 512-pixel patches to maintain spatial context while enabling detailed analysis. From this process, we generated a dataset of 2,135 triplets, each containing a pre-disaster image, a post-disaster image, and a manually annotated damage categorical mask.

    \begin{figure}[!htbp]
        \centering
        \includegraphics[width=\linewidth]{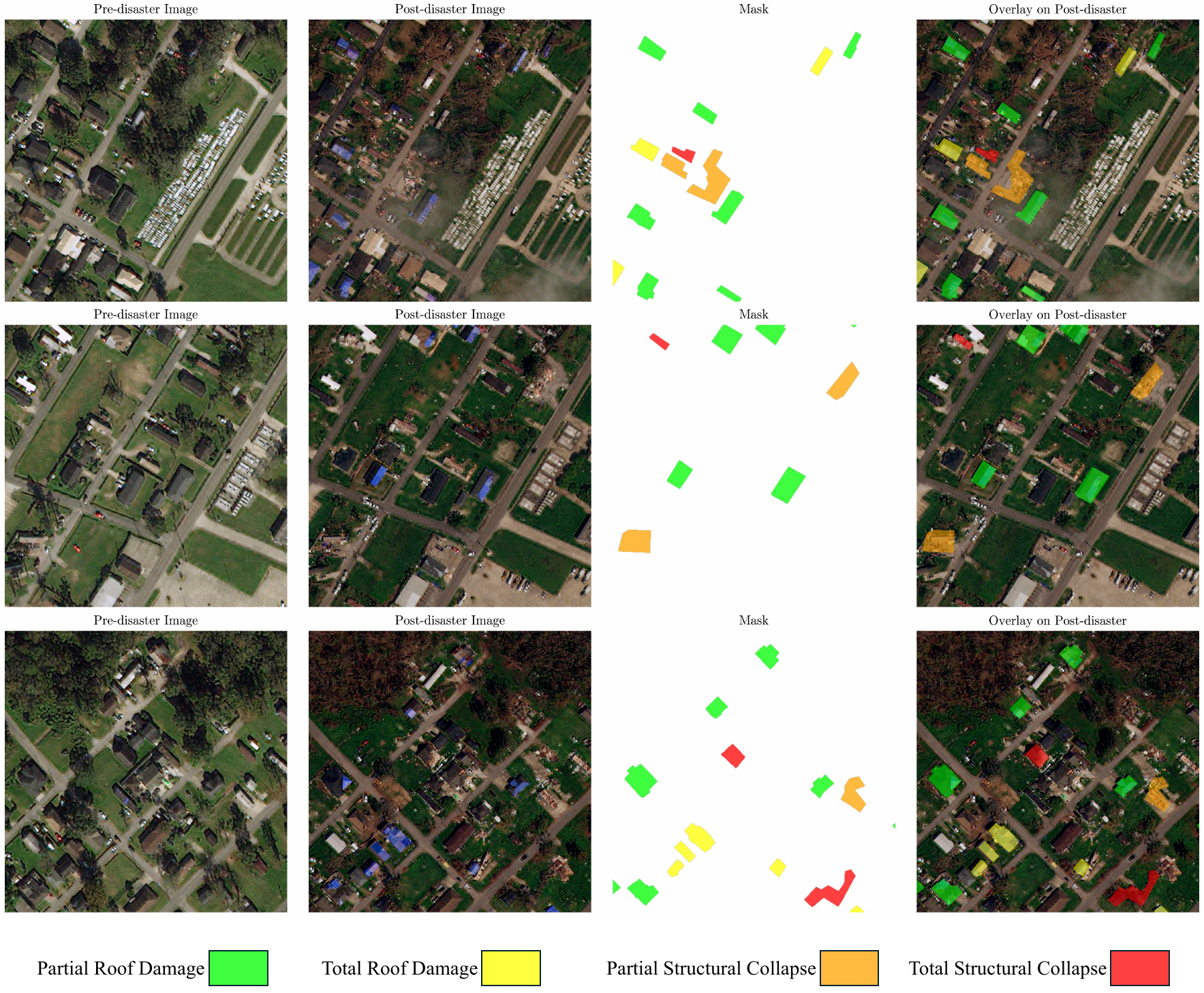}
        \caption{Example images with manually labeled annotations; the legend of the four damage types are shown below the picture.}\label{fig:anno}
        
    \end{figure}

Table~\ref{tab:anno_rules} details the specific annotation criteria used to define the four building damage categories within the BD-TypoSAT dataset. For each class—Partial Roof Damage, Total Roof Damage, Partial Structural Collapse, and Total Structural Collapse—the table provides both a structural definition describing the physical state of the building and a list of corresponding satellite cues. These cues are the visual indicators observable from top-down satellite imagery that annotators used to assign a building to a specific damage category. This systematic approach ensures consistency in labeling across the dataset and provides a clear framework for understanding the distinctions between the typological damage classes.

\begin{table}[ht]
    \centering
    \caption{Annotation criteria for BD-TypoSAT damage classes}\label{tab:anno_rules}
    
    \setlength{\tabcolsep}{4pt}     
    \renewcommand{\arraystretch}{1.15} 
  
    \begin{tabularx}{\linewidth}{@{}lLL@{}}   
      \toprule
      \textbf{Class} & \textbf{Structural definition} & \textbf{Satellite cues} \\
      \midrule
      Partial roof damage      &
        $\le{}50\,\%$ roof covering removed; walls intact &
        Irregular dark patches; rafters visible; footprint unchanged \\[2pt]
  
      Total roof damage        &
        $>50\,\%$ roof missing or complete deck loss; walls intact &
        Entire interior exposed; clean rectangular hole; debris nearby \\[2pt]
  
      Partial structural collapse &
        $\le{}50\,\%$ load-bearing walls failed &
        One corner flattened; L-shaped debris; shadows from remaining walls \\[2pt]
  
      Total structural collapse &
        $>50\,\%$ structural loss &
        Rubble mound; indistinct footprint \\
      \bottomrule
    \end{tabularx}
  \end{table}

   In this study, building damage is categorized into four specific classes---partial roof damage, total roof damage, partial structural collapse, and total structural collapse---thereby capturing key structural distinctions that go beyond coarse severity labels. Unlike many comparable datasets, our annotation exclusively focuses on damaged structures, deliberately omitting undamaged buildings to ensure that all labeled instances represent actual disaster-induced harm. As illustrated in Figure~\ref{fig:anno}, each damage type is color-coded for clarity, highlighting the diverse manifestations of structural failure. However, the distribution of annotated buildings skews toward less severe damage, making the rare instances of severe collapse comparatively underrepresented. To address this class imbalance during training, we adopt a composite loss function that fuses weighted cross-entropy with Dice loss, assigning each category a weight derived as the square root of its inverse frequency (see Table~\ref{tab:damage-classes}). By boosting the influence of scarce yet critical classes, this weighting strategy ensures that the model remains sensitive to all damage types, including those often overshadowed by more prevalent, lower-severity categories.

   \begin{table}[ht]
    \centering
    \caption{Distribution of building damage categories and corresponding frequency weights.}\label{tab:damage-classes}
    \begin{tabular}{
      @{}l
      S[table-format=4,round-mode=places,round-precision=0]
      S[table-format=2.2,table-space-text-post=\%]
      S[table-format=2.2]@{}
    }
      \toprule
      {Damage category} & {Count} & {Share (\%)} & {Weight ($\sqrt{1/f}$)}\\
      \midrule
      Partial roof damage        & 7030 & 86.69 & 1.07\\
      Total roof damage          &  958 & 11.81 & 2.91\\
      Partial structural collapse&   83 &  1.02 & 9.90\\
      Total structural collapse  &   38 &  0.47 & 14.61\\
      \bottomrule
    \end{tabular}
  \end{table}

The model employs a composite loss function formulated as:
\begin{equation}
\label{eq1}
Loss = \sum_{n \in C} w_{n}(0.5 \cdot dice_{n} + 0.5 \cdot ce_{n})
\end{equation}
where \(C\) represents the set of all categories, \(w_{n}\) denotes the weight assigned to the \(n\)-th category, \(dice_{n}\) is the dice loss, and \(ce_{n}\) is the cross entropy loss for category \(n\). Leveraging both Dice loss and cross-entropy in a composite loss function addresses the respective limitations of using either metric in isolation. Dice loss places greater emphasis on the spatial overlap between predicted and ground-truth masks, making it particularly effective in highly imbalanced segmentation tasks where underrepresented classes might otherwise be neglected. However, Dice loss can suffer from unstable gradients when the overlap is minimal---especially in early training or in scenarios with rare damage categories. In contrast, cross-entropy excels at pixel-wise classification accuracy, but it can struggle to account for severe class imbalances, potentially skewing model performance toward the majority class. By combining these two losses, we harness Dice loss's ability to robustly guide the segmentation of underrepresented classes while benefiting from cross-entropy's stable gradient flow and accurate pixel-level learning. This unified approach thus ensures balanced recognition of each damage category, from minor roof impairments to total structural collapse.

\subsection{Model Architecture}

\begin{figure}[!htbp]
    \centering
    \includegraphics[width=\linewidth]{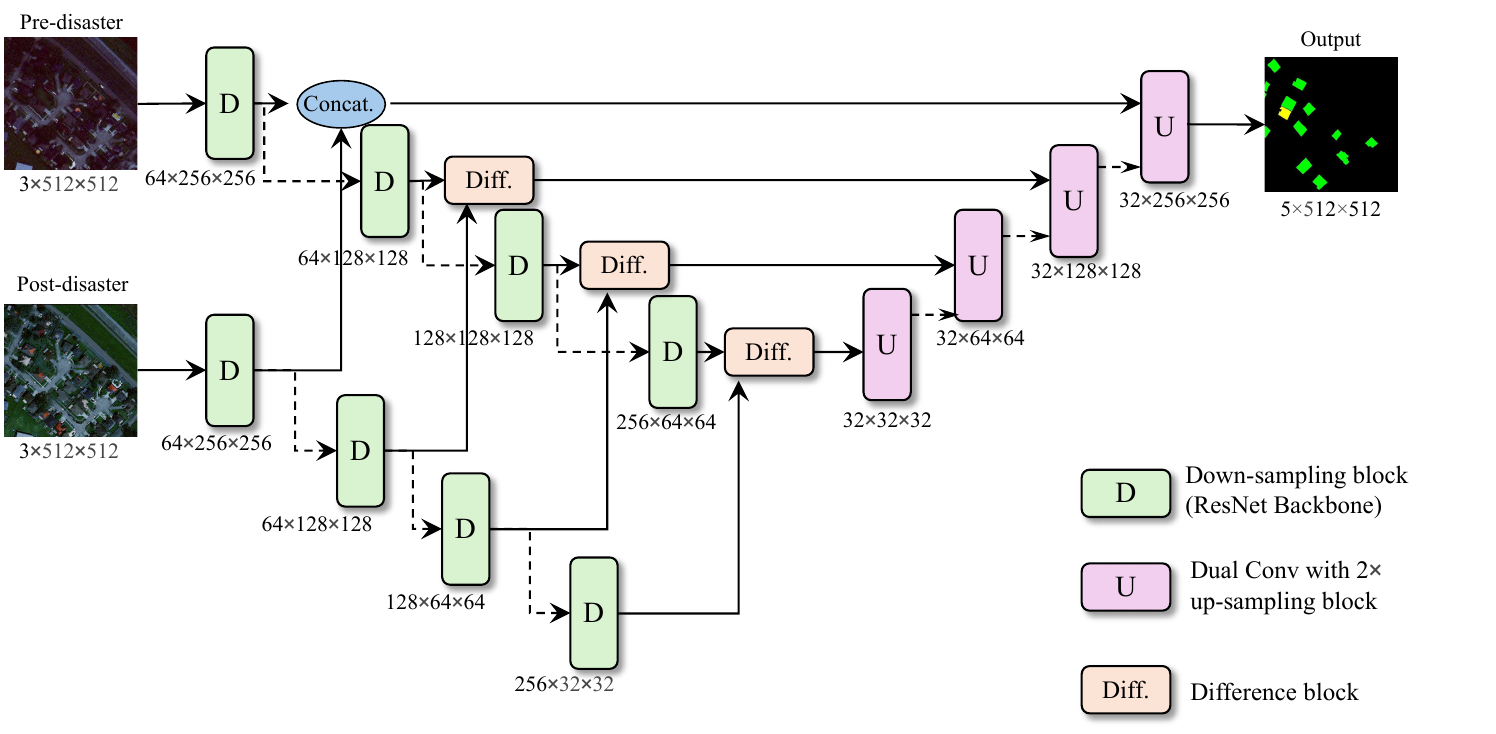}
    \caption{Transformer and U-Net-based architecture for disaster change detection and segmentation}\label{fig:arch}
    
\end{figure}

Figure~\ref{fig:arch} presents the DAHiTrA model we are using in this work, which is proposed by \citet{kaur_largescale_2023}. This hierarchical U-Net architecture is designed for satellite imagery change detection in disaster scenarios. The network processes pre-disaster and post-disaster images (\(3 \times 512 \times 512\)) through parallel ResNet-based down-sampling paths that progressively extract hierarchical features at decreasing resolutions. Each green block represents a down-sampling stage that halves spatial dimensions while increasing feature depth, creating multi-scale feature representations from \(64 \times 256 \times 256\) down to \(256 \times 32 \times 32\). At three intermediate levels, the difference blocks compute and analyze changes between corresponding feature maps from both temporal states. These difference blocks capture change information at multiple scales, from fine details to broader contextual changes. The decoders (purple blocks) include a dual convolutional block with intermediate normalization and \(2\times\) up-sampling operation to gradually restore spatial resolution while integrating change information from multiple scales. Skip connections (dashed lines) provide direct pathways for high-resolution spatial information to flow into the decoder. The network ends with a concatenation operation that merges the highest-level features with the original resolution features. This creates a final \(5 \times 512 \times 512\) output that classifies each pixel into five distinct change categories (one of which is the background). This architecture effectively balances fine-grained spatial detail with hierarchical feature representation, making it suitable for identifying structural changes in post-disaster environments.

\subsection{Experimental Setup}

All experiments were conducted on a workstation equipped with an NVIDIA RTX 6000 GPU with 48GB of VRAM.
We trained our model using the following hyperparameters: learning rate of 0.001 with a linear decay policy, input image size of 512 \(\times\) 512 pixels, batch size of 8, and a maximum of 300 epochs. The Adam optimizer was used with default \(\beta\) values (\(\beta_1 = 0.9\), \(\beta_2 = 0.999\)) and \(\epsilon = 10^{-8}\). Early stopping was implemented with a patience parameter of 20 epochs to prevent overfitting, and the model weights from the epoch with the lowest validation loss were selected for evaluation.
We implemented a custom data augmentation pipeline to increase the effective size of the dataset. Images were processed at 512 \(\times\) 512 pixels and normalized with mean and standard deviation values of [0.5, 0.5, 0.5]. The dataset was split into training and validation sets with an 90:10 ratio to ensure robust model evaluation while maintaining sufficient training examples. Prior to train-validation split, a test set of 10\% (213 datapoints) of the entire dataset was held-out for both qualitative and quantitative evaluation. The training was conducted four times, with different random seed set when splitting train-validaion-test sets to ensure robustness. Our augmentation strategy included random horizontal and vertical flipping, 90$^{\circ}$/180$^{\circ}$/270$^{\circ}$ rotations, and Gaussian blur with random radius. During training, we employed a patch-based cropping approach where random crops were extracted from training images, while validation used predefined patch positions. To address the class imbalance issue, we implemented an upsampling technique that duplicated training samples containing specific class labels (particularly those with positive values in classes 2--4), with additional duplication for samples containing classes 2 or 3, effectively increasing the representation of less frequent classes in the training dataset.
The model was implemented using PyTorch 2.5.1.

\section{Results}\label{sec4}

\subsection{Quantitative Evaluation}

\begin{table}[ht]
\centering
\caption{Validation-set performance averaged over four random data splits (\(n=4\)).  
Values are reported as mean \(\pm\) standard deviation.}
\label{tab:val_metrics_summary}
\begin{tabular}{lcccc}
\toprule
\textbf{Category} & \textbf{F1} & \textbf{IoU} & \textbf{Precision} & \textbf{Recall}\\
\midrule
Partial roof damage            & \(0.809\pm0.008\) & \(0.680\pm0.011\) & \(0.860\pm0.013\) & \(0.785\pm0.009\)\\
Total roof damage              & \(0.838\pm0.014\) & \(0.722\pm0.021\) & \(0.917\pm0.010\) & \(0.792\pm0.032\)\\
Partial structural collapse    & \(0.900\pm0.056\) & \(0.822\pm0.088\) & \(0.960\pm0.012\) & \(0.865\pm0.093\)\\
Total structural collapse      & \(0.842\pm0.061\) & \(0.732\pm0.092\) & \(0.966\pm0.012\) & \(0.781\pm0.108\)\\
\midrule
\textbf{Overall (macro)}       & \(\mathbf{0.846\pm0.021}\) & \(\mathbf{0.737\pm0.033}\) & \(\mathbf{0.914\pm0.005}\) & \(\mathbf{0.799\pm0.038}\)\\
\bottomrule
\end{tabular}
\end{table}

\begin{figure}[htbp]
    \centering
    \includegraphics[width=\textwidth]{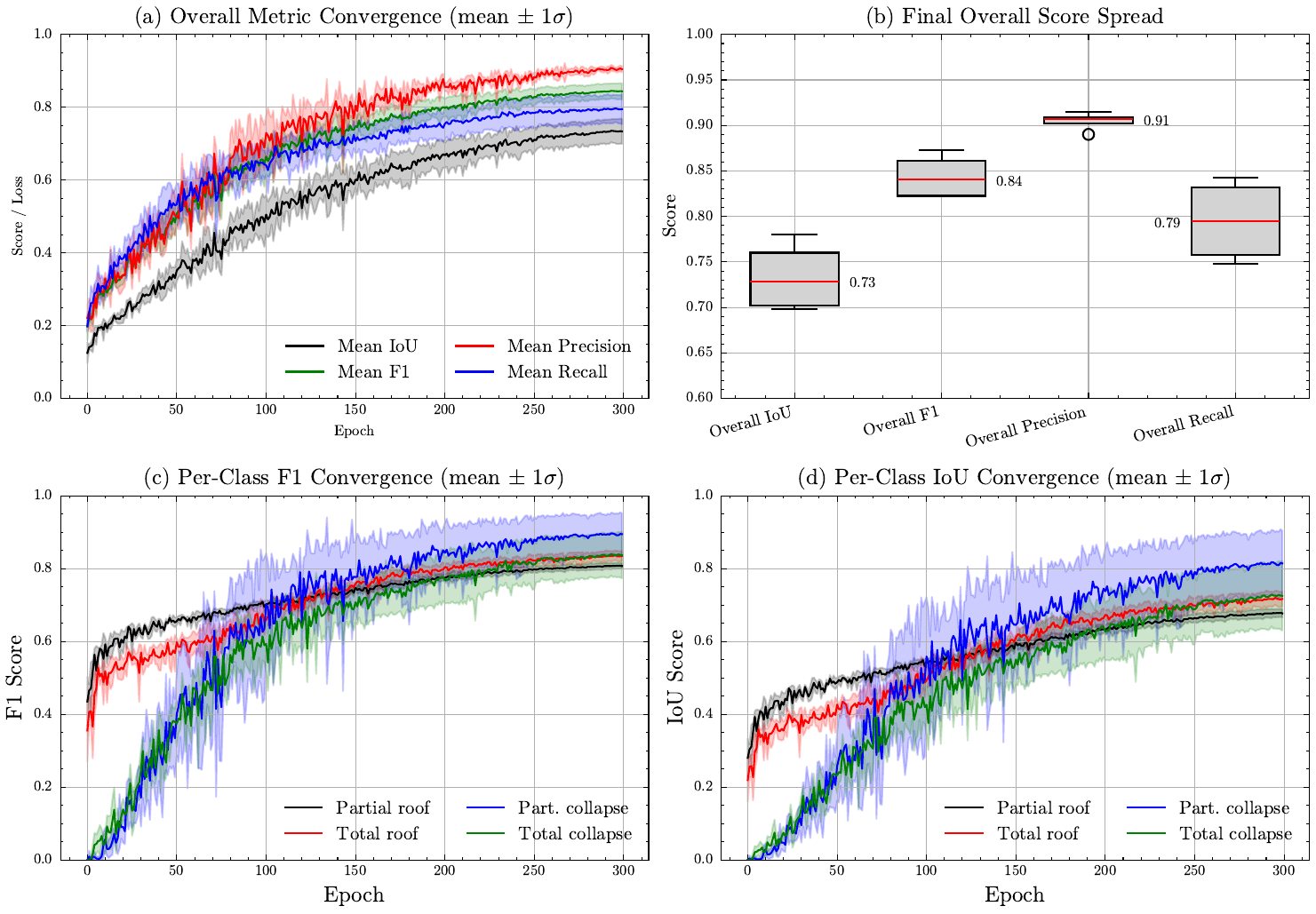}
    \caption{Convergence and variance of validation metrics.\ (a) shows the mean learning curves for IoU, F1 score, precision, and recall across all runs, with the shaded region representing \(\pm \sigma\).\ (b) shows the spread of final epoch scores for these metrics using box plots.\ (c) and (d) show the learning curves of mIoU and F-1 scores for the four classes across the four runs.}\label{fig:convergence_variance}
    
\end{figure}

The model was trained for the full 300 epochs as specified in our experimental setup. We monitored several key validation metrics throughout the training process. Figure~\ref{fig:convergence_variance} illustrates the progression and distribution of various performance metrics across the four independent training runs. The complete breakdown of per-class metrics (IoU, F1, precision, recall) for each training run and their progression over epochs are presented in Figures~\ref{fig:app_val_iou_f1_barcharts}--\ref{fig:app_val_log_metrics_run4} in Appendix~\ref{sec:appendix_figures}.

Figure~\ref{fig:convergence_variance} (a) displays the convergence of overall validation metrics (Mean IoU, Mean F1, Mean Precision, and Mean Recall), averaged across the runs. These metrics generally exhibit a consistent upward trajectory, stabilizing as training progresses, indicating effective learning. The shaded regions represent one standard deviation across runs, highlighting the variability in performance. As summarized in Table~\ref{tab:val_metrics_summary}, the model achieves a peak overall macro F1-score of \(0.846\pm0.021\) and a macro IoU of \(0.737\pm0.033\).
Figure~\ref{fig:convergence_variance} (b) shows box plots that summarize the distribution of the overall IoU, F1, Precision, and Recall scores specifically from the final training epoch of each of the four runs. This subplot illustrates the consistency and spread of the model's overall performance at the conclusion of the fixed 300-epoch training regime.

The convergence patterns of per-class F1 scores and IoU scores are detailed in Figures~\ref{fig:convergence_variance} (c) and (d), respectively. These plots show the mean and standard deviation across runs for four key damage categories: ``Partial roof damage'', ``Total roof damage'', ``Partial structural collapse'', and ``Total structural collapse'', over the 300 epochs. It is evident that class-specific disparities in learning speed and stability are most pronounced in the earlier epochs (e.g., before epoch 75).

Referring to Table~\ref{tab:val_metrics_summary} for the peak performance of each class (averaged best scores from four runs):
The ``Partial structural collapse'' class achieved the highest F1-score (\(0.900\pm0.056\)) and IoU (\(0.822\pm0.088\)). This strong, albeit more variable (as indicated by the standard deviation and visible in early convergence in Fig.~\ref{fig:convergence_variance}c,d) outcome may be attributed to its relatively distinctive visual cues.

The ``Total roof damage'' category also performed well, with an F1-score of \(0.838\pm0.014\) and IoU of \(0.722\pm0.021\). ``Partial roof damage'' yielded an F1 of \(0.809\pm0.008\) and IoU of \(0.680\pm0.011\). The performance on these roof categories, despite their visual subtleties and higher instance counts, is robust.
The ``Total structural collapse'' class achieved an F1-score of \(0.842\pm0.061\) and IoU of \(0.732\pm0.092\). Its performance, particularly the higher variability in recall (\(0.781\pm0.108\)), might be influenced by its lower representation in the dataset (38 labeled examples). Detailed per-class, per-run bar charts illustrating the spread of these best scores are available in Appendix~\ref{sec:appendix_figures}.

Despite the inherent challenges of class imbalance and visual similarity in satellite imagery, the model demonstrates strong performance. The convergence trends in Figure~\ref{fig:convergence_variance}, coupled with robust F1 scores across all damage categories (ranging from \(0.809\) to \(0.900\), as per Table~\ref{tab:val_metrics_summary}), highlight the effectiveness of our modeling choices—including the difference blocks, transformer-based architecture, and balanced loss function. While strategies like weighted loss and data augmentation helped mitigate imbalance, the results reveal areas for potential improvement, such as acquiring additional targeted samples for less represented classes.

To contextualize our model's performance, we compare its overall macro F1-score with those reported in several recent studies on building damage assessment using satellite imagery. Table~\ref{tab:comparison_sota_transposed} presents this comparison with other methods, many of which focus on binary or ordinal damage classification. It is important to note that direct comparisons can be nuanced due to differences in datasets, evaluation protocols, and the specific definitions of damage categories used in each study. Consequently, we do not report per-class metrics: the typology-based damage classes introduced in DamageCAT share no one-to-one correspondence with the binary or ordinal labels used in prior work, so a class-level comparison would be conceptually misleading.

\begin{table}[ht]
\centering
\caption{Overall F1-score comparison between our model and prior work}\label{tab:comparison_sota_transposed}

\begin{tabular}{p{7cm}c}
\toprule
\textbf{Method} & \textbf{F1 Score} \\
\midrule
UNet{+}BDANet {+}morphology, \citet{neto_building_2024} & 0.7998 \\
MTF,~\citet{weber_building_2020} & 0.7417 \\
HRTBDA,~\citet{chen_hrtbda_2024} & 0.8071 \\
\citet{bouchard_transfer_2022} & 0.8460 \\
\citet{wang_building_2022} & 0.8322 \\
\citet{liu_novel_2022} & 0.7017 \\
\midrule
\textbf{Ours} & \textbf{0.8458} \\
\bottomrule
\end{tabular}
\end{table}

\subsection{Qualitative Analysis}

\begin{figure}[htbp]
    \centering
    \includegraphics[width=\linewidth]{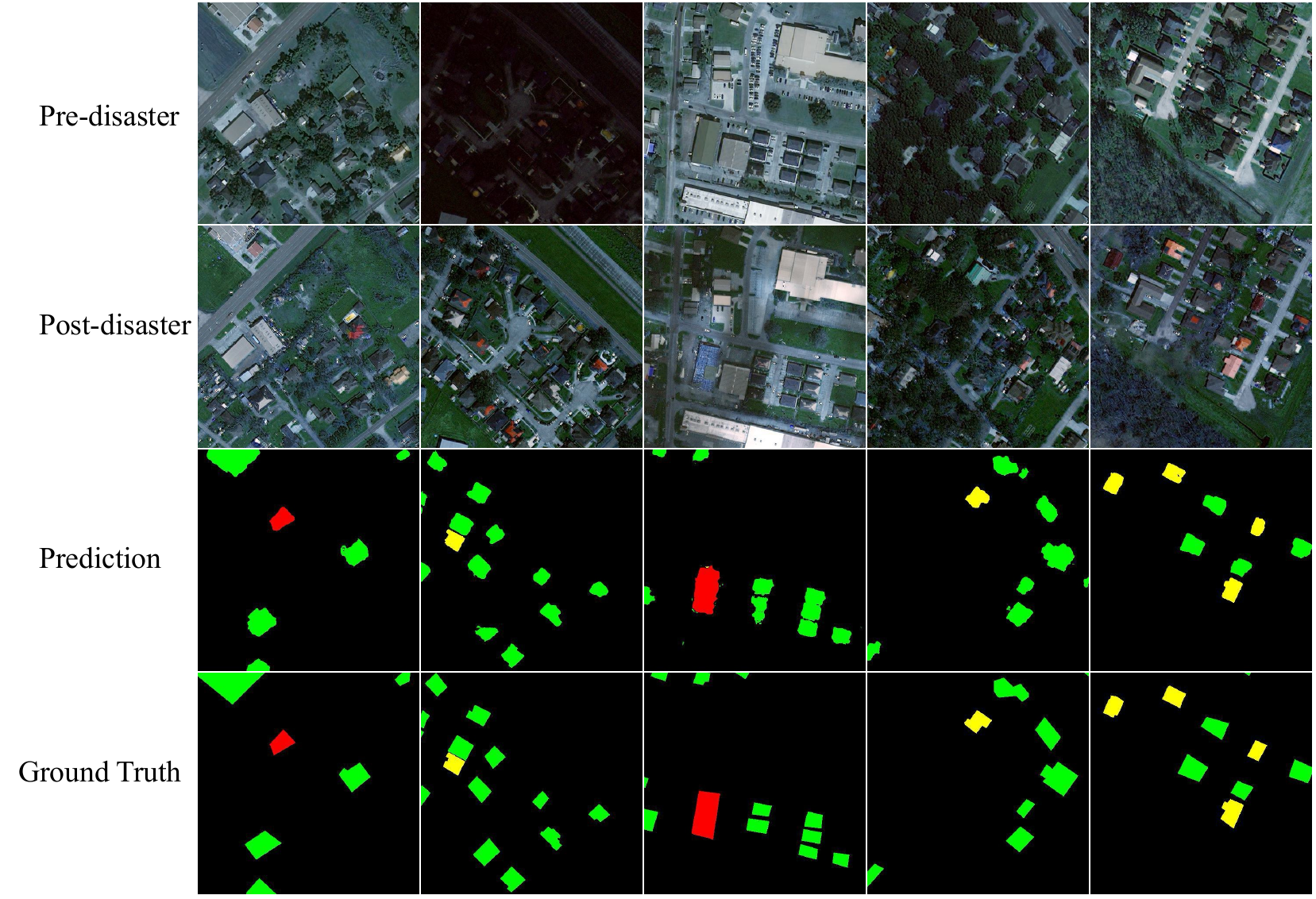}
    \caption{The qualitative evaluation result of 5 select image data points.}\label{fig:eval}
    
\end{figure}

To supplement the quantitative metrics, we conducted a visual evaluation of model performance using five representative image samples, selected to encompass diverse damage scenarios and varied building typologies (Figure~\ref{fig:eval}). Each sample contains pre- and post-disaster satellite views alongside the model's prediction mask and ground truth annotations, offering a side-by-side comparison of detection and classification outcomes.

This qualitative examination shows general agreement between predicted masks and annotated references. This suggests the model can distinguish multiple damage categories. Even in dense urban areas, where many damaged structures are close together, the model appears to delineate individual building boundaries and assign damage labels with reasonable accuracy. This indicates an ability for spatial discrimination. Notably, the model consistently recognizes differences between partial roof damage and partial structural collapse in adjacent buildings. This suggests it can learn some visual cues associated with each damage type.

Overall, these visual findings demonstrate the model's practical utility in emergency response contexts, where rapid, detailed mapping of post-disaster impacts is essential. Unlike binary or severity-only classification methods, this approach delivers categorical insights---pinpointing whether a roof is simply compromised or an entire structure is at risk of collapse. Such richer information could be valuable for guiding targeted interventions and resource allocation. Moreover, the consistency of these results across varied scenarios suggests some robustness in our typology-based system. This is important because defining damages with such specific detail is a novel aspect of our system, and direct benchmarks are not available in current literature.

\begin{figure}[htbp]
    \centering
    \includegraphics[width=\linewidth]{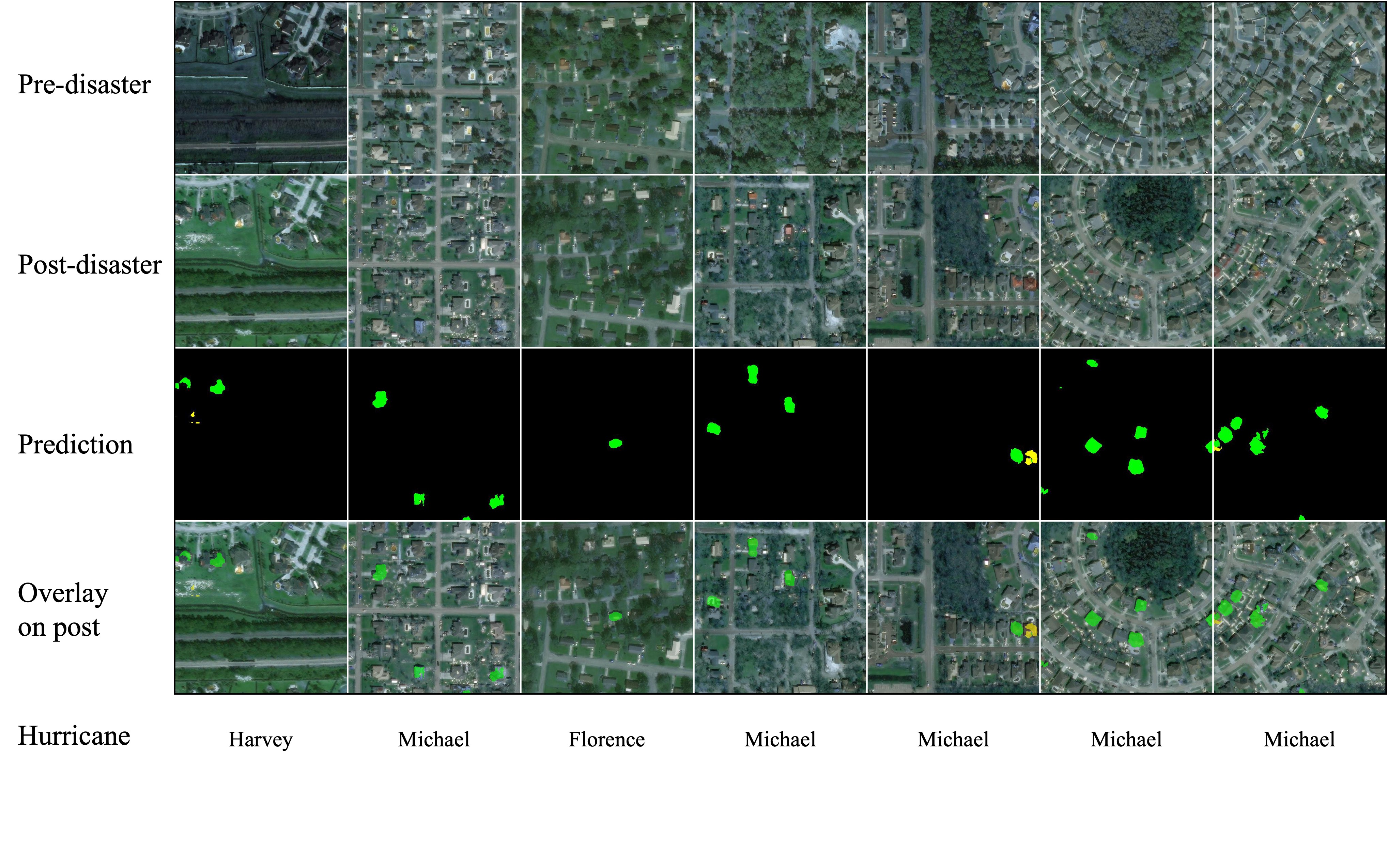}
    \caption{Model transferability evaluation across multiple hurricane events. The figure shows prediction results on Hurricane Harvey (column 1), Hurricane Florence (column 3), and Hurricane Michael (columns 2, 4--7). Each row displays pre-disaster imagery, post-disaster imagery, model predictions, and overlay visualization respectively.}\label{fig:eval_multi}
    
\end{figure}

To evaluate the model's transferability beyond Hurricane Ida, we tested it on satellite images from three other hurricanes: Harvey (2017), Florence (2018), and Michael (2018), as shown in Figure~\ref{fig:eval_multi}, which all come from the dataset xBD \citep{gupta_xbd_2019}. The results demonstrate the model's ability to generalize across different hurricane events, locations, and types of buildings without needing extra training. The model successfully found damaged buildings in the Houston area from Hurricane Harvey (leftmost column) and in North Carolina's coastal areas after Hurricane Florence (column 3). The results from Hurricane Michael (columns 2 and 4--7) also show it worked well across Florida's Panhandle region. Although the model's confidence sometimes varied, likely due to factors like image quality, different seasons, and local building styles, it consistently classified the type of damage correctly across all these hurricanes. This consistency suggests our framework could be a valuable tool in real-world disaster response. However, quantitative evaluation of these cross-disaster results was not feasible due to fundamental differences in labeling criteria between our typology-based damage categories and the severity-based annotations employed in the xBD dataset.

\section{Discussion}\label{sec:5}

\subsection{Interpretation of results}
Through experimental validation, we establish that a categorical, deep-learning-driven approach to building damage assessment can effectively leverage pre- and post-disaster satellite imagery. The performance metrics for the four damage categories provide important insights into the method's strengths and remaining challenges. Notably, the rare ``partial structural collapse'' class achieved unexpectedly strong results (IoU \(\approx 0.8344\), F1 \(\approx 0.9097\)), surpassing metrics for the more common ``partial roof damage'' category (IoU \(\approx 0.7420\), F1 \(\approx 0.8519\)). This discrepancy highlights that visual distinctiveness (e.g., exposed interior walls or collapsed debris fields) can override class imbalance constraints, making certain damage patterns more readily identifiable from a top-down viewpoint.

However, persistent performance gaps across categories---even after applying weighted losses and data augmentation---emphasize the intrinsic difficulties in using vertical satellite imagery for detailed damage classification. Roof damage, in particular, presents subtle visual cues (e.g., missing shingles or partial roof tarps) that can be masked by factors like shadowing, overhanging vegetation, or variations in roof color. Consequently, the two roof damage classes exhibit relatively lower IoU and F1 scores compared to structural collapse classes, whose larger-scale changes (e.g., caved-in walls, rubble) are more conspicuous in vertical imagery. Even though the 0.4~m/pixel resolution images provided valuable detail, fine-grained features remain challenging to discern due to atmospheric interference, illumination conditions, and satellite angle variations---an issue corroborated by \citet{chen_hrtbda_2024} and \citet{liu_novel_2022}.

Overall, these outcomes underscore the need for continued innovation, such as multi-modal data integration or oblique and street-level perspectives, to capture more subtle roof-related damage patterns. Nonetheless, the robust performance observed across all categories confirms the promise of our typology-based approach, offering a nuanced alternative to traditional severity-focused models and paving the way for more targeted disaster response and recovery strategies.

While our typology-based approach represents an advancement in damage classification granularity, it builds upon decades of prior work in remote sensing-based damage assessment. Earlier studies have attempted various forms of detailed damage categorization, though often constrained by the resolution limits of available imagery and computational capabilities of their time. Our framework's performance, while promising, still faces some of the fundamental challenges identified in seminal works: the inherent limitations of vertical imagery for detecting certain damage types, variations in classification accuracy across different urban morphologies, and the persistent challenge of rare damage cases. These constraints underscore that while deep learning enables more nuanced damage classification, careful consideration of operational contexts and potential accuracy trade-offs remains essential for practical deployment.

\subsection{Significance and implications}

The DamageCAT framework advances building damage assessment by moving beyond coarse severity ratings (e.g., ``minor,'' ``moderate,'' ``major'') to typology-based classifications that specify the nature and location of structural damage. This approach addresses a key research gap identified by \citet{melamed_uncovering_2024}, enabling transition from generic damage labels to well-defined categories such as partial roof damage versus total structural collapse. Methodologically, our transformer-enhanced U-Net architecture effectively isolates disaster-induced changes by comparing pre- and post-disaster satellite imagery, filtering out pre-existing conditions as advocated by \citet{zheng_building_2021} and \citet{ahn_generalizable_2025}. The multi-scale difference blocks enable robust detection of subtle structural variations even in highly imbalanced datasets.

The practical implications for emergency response are significant. Traditional severity classifications provide limited actionable information, whereas typological damage assessment enables targeted resource allocation. For example, buildings with partial roof damage require roofing specialists and materials, while those with total structural collapse need heavy machinery and search-and-rescue teams. This specificity reduces operational uncertainties and accelerates response times compared to generic severity ratings.

Beyond immediate response, typological classification enhances long-term recovery planning and risk assessment. Engineers can prioritize critical repairs based on specific damage modes, while insurers gain more precise loss estimates. City planners can identify patterns of structural vulnerability to inform future building codes and disaster preparedness strategies. The framework's granular damage descriptions also improve transparency in post-disaster decision-making, providing clear, consistent information for stakeholders ranging from first responders to policy makers.

\subsection{Challenges and limitations}

Despite promising results, several key limitations constrain the framework's immediate scalability and generalizability. These challenges span data acquisition, methodological constraints, and practical deployment considerations.

Data acquisition and costs represent the primary barrier to widespread adoption. While building-level damage assessment requires sub-meter resolution imagery (\(\sim\)0.5~m/pixel), most freely available sources like Sentinel-1/2 and Landsat provide only 10--30~m resolution---adequate for flood mapping but insufficient for distinguishing roof damage from structural collapse. Commercial alternatives are costly: as shown in Table~\ref{tab:imagery_costs}, high-resolution imagery ranges from \$2.25/km\(^2\) for 3~m archives to \$22.50/km\(^2\) for 30~cm data, with ultra-high-resolution SAR costing \$675--\$3{,}250 per 5~km \(\times\) 5~km scene. Even when emergency providers like Maxar Open Data or Planet Labs donate post-disaster imagery, comprehensive pre-disaster coverage typically requires commercial acquisition. Manual annotation compounds these costs, particularly for subtle damage distinctions that demand expert interpretation.

Methodological limitations stem from relying solely on nadir satellite imagery, which can miss facade-level or internal damage visible only from oblique angles. Severe class imbalance—with total structural collapse representing just 0.47\% of annotated buildings—affects model robustness despite weighted loss mitigation. The framework's training on a single event (Hurricane Ida) in one region (New Orleans, Louisiana) also limits generalizability, as damage patterns vary significantly across hazard types and building typologies \citep{adriano_developing_2023}.

Computational requirements from the transformer architecture may hinder real-time deployment in resource-constrained environments. During crises, inference speed often matters more than marginal accuracy gains \citep{gholami_deployment_2022}, suggesting the need for model optimization techniques like distillation or quantization.

Addressing these limitations through multi-modal data integration, lighter model architectures, and extensive cross-regional validation represents an important avenue for future research.

\begin{table}[ht]
\centering
\caption{Indicative costs and access models for satellite imagery in disaster response.}\label{tab:imagery_costs}

\begin{tabular}{p{4cm}p{2.9cm}p{5.4cm}p{1cm}}
\toprule
\textbf{Provider / Product} & \textbf{Native resolution} & \textbf{Typical cost or licence} & \textbf{Source}\\
\midrule
PlanetScope (archive) & 3~m & \$2.25/km\(^2\) (min 250~km\(^2\)) &\cite{apollo_planetscope}\\
Pléiades Neo (archive) & 30~cm & \$22.50/km\(^2\) (min 25~km\(^2\)) &\cite{apollo_pleiades}\\
\textbf{Maxar Open Data} & 30--50~cm & Free for sudden-onset disasters &\cite{maxar_open_data}\\
Planet Labs & 3--5~m & Free on request for accredited responders &\cite{planet_disaster}\\
Umbra SAR tasking & 0.25--1~m radar & \$675--\$3{,}250 per 5~km \(\times\) 5~km scene &\cite{umbra_pricing}\\
\mbox{Copernicus EMS} & 0.5--10~m (mixed) & Free maps; imagery cost covered by EU &\cite{copernicus_ems}\\
Sentinel 1/2 & 10--20~m optical/radar & Free open access &\cite{copernicus_access}\\
Landsat archive & 15--30~m & Free open access &\cite{usgs_free_landsat}\\
International Charter & Mixed sensors & Free emergency tasking (authorised users) &\cite{intl_charter}\\
\bottomrule
\end{tabular}
\end{table}

\subsection{Future research directions}

Several research directions emerge from this work's limitations and findings. Dataset expansion represents the most critical need, requiring balanced representation across damage categories, disaster types, and geographic regions to improve model generalizability. Current class imbalances and regional specificity limit broader applicability, suggesting the need for coordinated data collection efforts across multiple hazard events.

Multi-modal data integration offers significant potential for improving classification accuracy. Incorporating oblique aerial photography and street-view imagery could provide complementary perspectives on building damage that nadir satellite imagery cannot capture \citep{braik_automated_2024, xue_post-hurricane_2024}. Additionally, integrating multiple post-disaster time points would enable tracking of recovery progression and damage evolution over time.

Model interpretability and optimization represent important technical directions. Developing interpretable models that highlight specific visual features contributing to damage classifications would enhance trust and adoption among emergency responders. Simultaneously, optimizing architectures for edge devices with limited computational resources is essential for field deployment in resource-constrained environments.

Operational integration requires interdisciplinary collaboration between computer scientists, disaster researchers, and emergency management practitioners. Future systems should prioritize actionable intelligence delivery, ensuring that technical advances translate into improved disaster response capabilities. This includes developing standardized evaluation protocols and establishing partnerships with operational agencies to validate real-world utility.

These research directions collectively aim to transform typological damage assessment from a promising research concept into a robust operational tool for disaster response and recovery planning.

\section{Conclusion}\label{sec:6}

This study introduced DamageCAT, a framework for typology-based building damage assessment that advances beyond traditional severity-focused methods. We contribute the BD-TypoSAT dataset, containing human-annotated satellite imagery with four distinct damage categories (partial roof damage, total roof damage, partial structural collapse, and total structural collapse), addressing the need for descriptive training data that captures the multifaceted nature of structural damage. Our hierarchical U-Net-based transformer architecture analyzes pre- and post-disaster satellite imagery pairs to produce detailed damage classification maps, achieving robust performance with 0.737 IoU and 0.846 F1-score across all damage types.

The framework's key advancement lies in shifting from broad severity categories to typology-based assessment, providing the detailed information necessary for effective resource allocation. By distinguishing roof damage from structural collapse, emergency responders can prioritize personnel, materials, and financial resources more strategically. The pre-post disaster change detection approach also reduces false positives by filtering out pre-existing building conditions, enhancing reliability for time-critical response situations.

Several noteworthy limitations should be acknowledged. Dataset imbalance, particularly for total structural collapse, reflects real-world damage distributions but indicates the need for expanded, balanced datasets. The reliance on nadir satellite imagery may obscure facade-level or internal damage features that oblique perspectives could reveal. Additionally, training on a single event (Hurricane Ida) in one region (New Orleans, Louisiana) limits generalizability, while the transformer architecture's computational demands may challenge real-time deployment in resource-constrained environments.

Future research should prioritize balanced data collection across multiple hazards and geographic regions, integration of multi-modal sources (oblique aerial imagery, street-view data, LiDAR), and model optimization for real-time inference. Developing interpretable AI methods could enhance user trust by revealing visual features driving classifications. Cross-regional validation and field testing will be essential for refining the typology-based paradigm and ensuring widespread adoption in operational disaster management contexts.

\section*{Declarations}

\bmhead{Acknowledgments}
The authors acknowledge the work of Annie Bone, Kaleb Valentin, and Kevin Rodriguez in annotating the dataset used in this study.

\bmhead{Funding}
This material is supported by the National Science Foundation under the CAREER grant (No. 1846069). Any opinions, findings, conclusions, or recommendations expressed in this material are those of the authors and do not necessarily reflect the views of the National Science Foundation.

\bmhead{Data and Code availability}
The data that support the findings of this study are publicly available at \href{https://zenodo.org/records/15453772}{Zenodo}, and the code is available at \href{https://github.com/YimingXiao98/DamageCAT}{GitHub}.

\newpage
\appendix\label{sec:appendix}

\section{Supplementary Figures}\label{sec:appendix_figures}

\begin{figure}[htbp]
    \centering
    \begin{subfigure}[b]{\textwidth}
        \includegraphics[width=\textwidth]{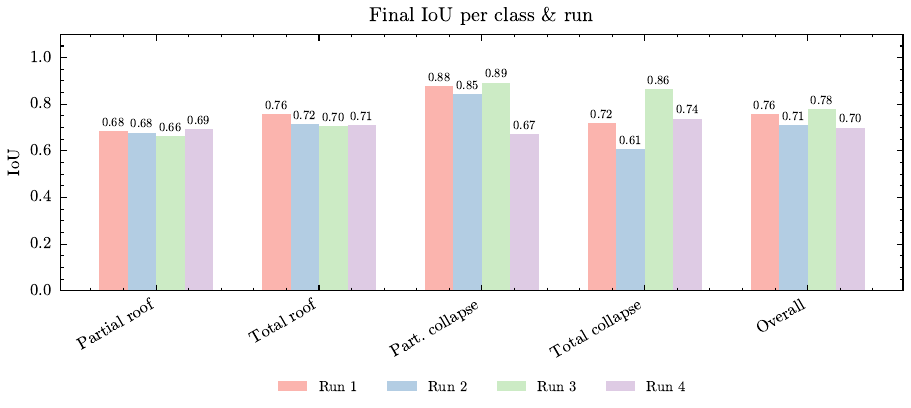}
        \caption{Per-Class IoU (Validation)}\label{fig:app_val_per_class_iou}
        
    \end{subfigure}
    \vspace{1em} 
    \begin{subfigure}[b]{\textwidth}
        \includegraphics[width=\textwidth]{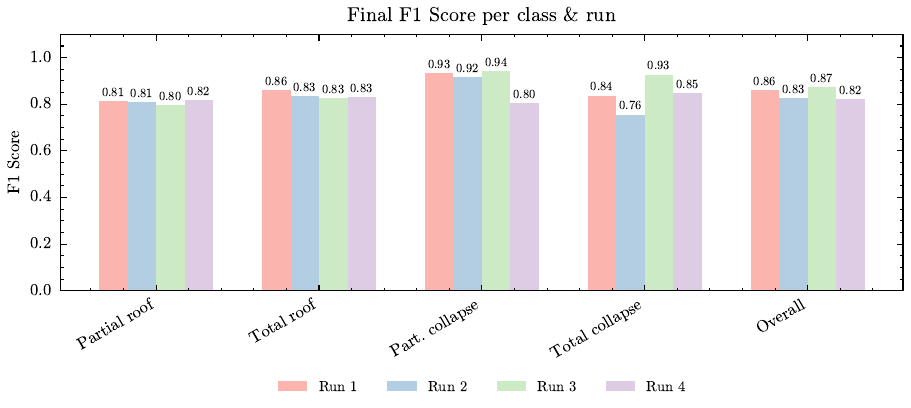}
        \caption{Per-Class F1-Score (Validation)}\label{fig:app_val_per_class_f1}
        
    \end{subfigure}
    \caption{Bar charts of (a) per-class IoU and (b) per-class F1-score, derived from validation metrics. For each class, bars represent the highest score achieved during training for each of the four runs, along with the average across runs. Error bars on the average indicate standard deviation.}\label{fig:app_val_iou_f1_barcharts}
    
\end{figure}

\begin{figure}[htbp]
    \centering
    \begin{subfigure}[b]{\textwidth}
        \includegraphics[width=\textwidth]{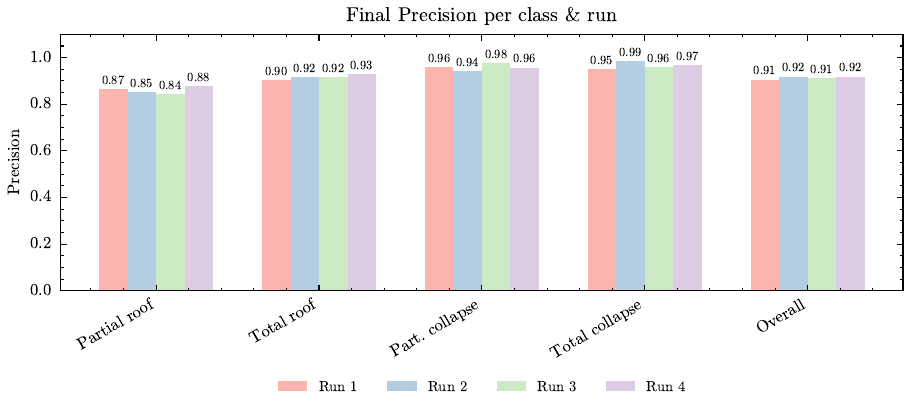}
        \caption{Per-Class Precision (Validation)}\label{fig:app_val_per_class_precision}
        
    \end{subfigure}
    \vspace{1em} 
    \begin{subfigure}[b]{\textwidth}
        \includegraphics[width=\textwidth]{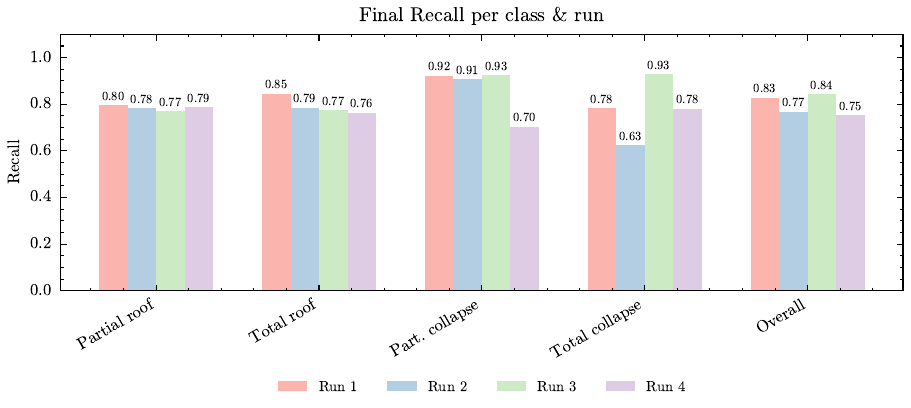}
        \caption{Per-Class Recall (Validation)}\label{fig:app_val_per_class_recall}
        
    \end{subfigure}
    \caption{Bar charts of (a) per-class Precision and (b) per-class Recall, derived from validation metrics. For each class, bars represent the highest score achieved during training for each of the four runs, along with the average across runs. Error bars on the average indicate standard deviation.}\label{fig:app_val_precision_recall_barcharts}
    
\end{figure}

\begin{figure}[htbp]
    \centering
    \includegraphics[width=\textwidth]{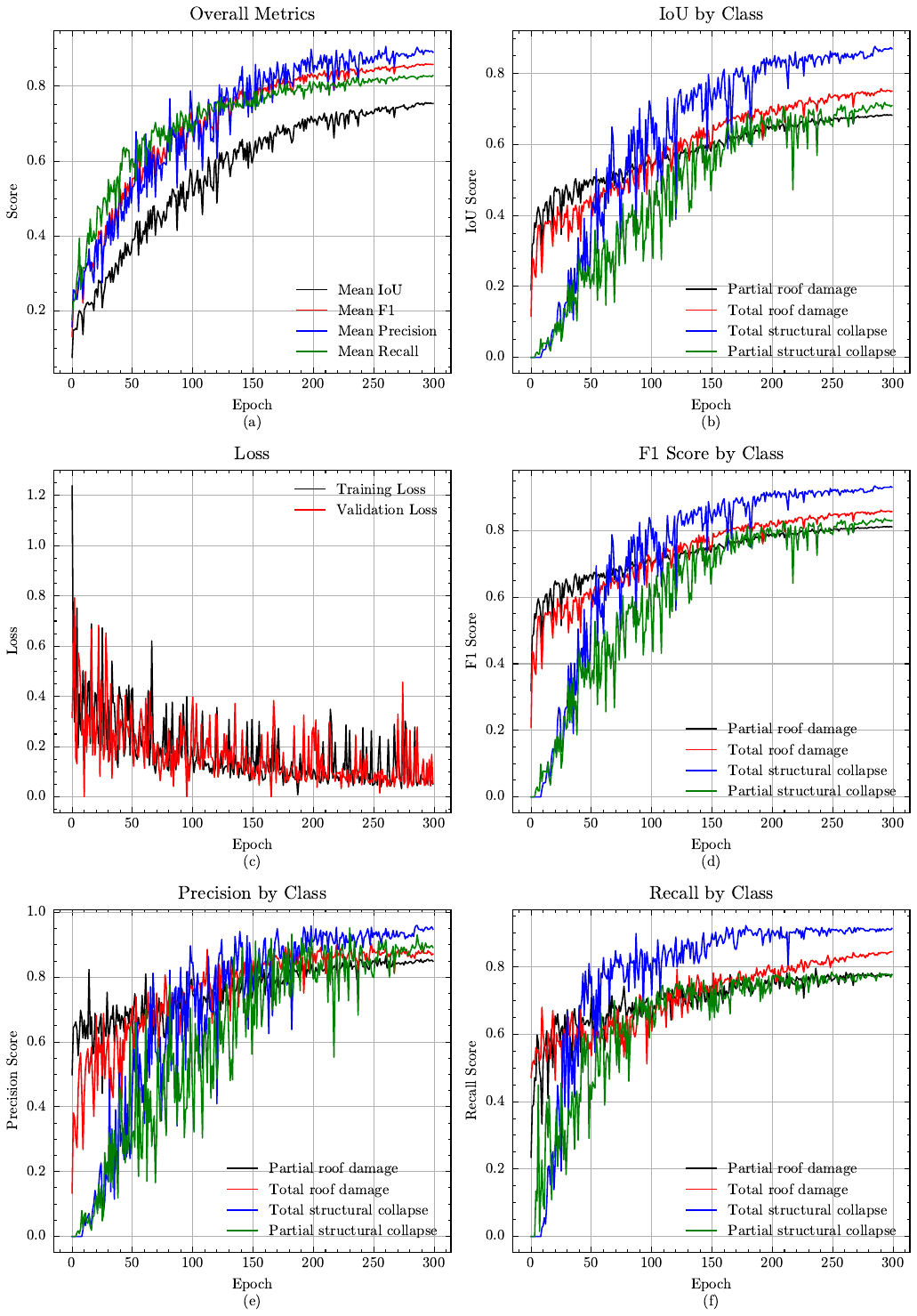}
    \caption{Progression of validation metrics throughout training epochs for Run 1. The plot displays: overall performance (IoU, F1, Precision, Recall), training vs. validation loss, and per-class IoU, F1-score, Precision, and Recall.}\label{fig:app_val_log_metrics_run1}
    
\end{figure}

\begin{figure}[htbp]
    \centering
    \includegraphics[width=\textwidth]{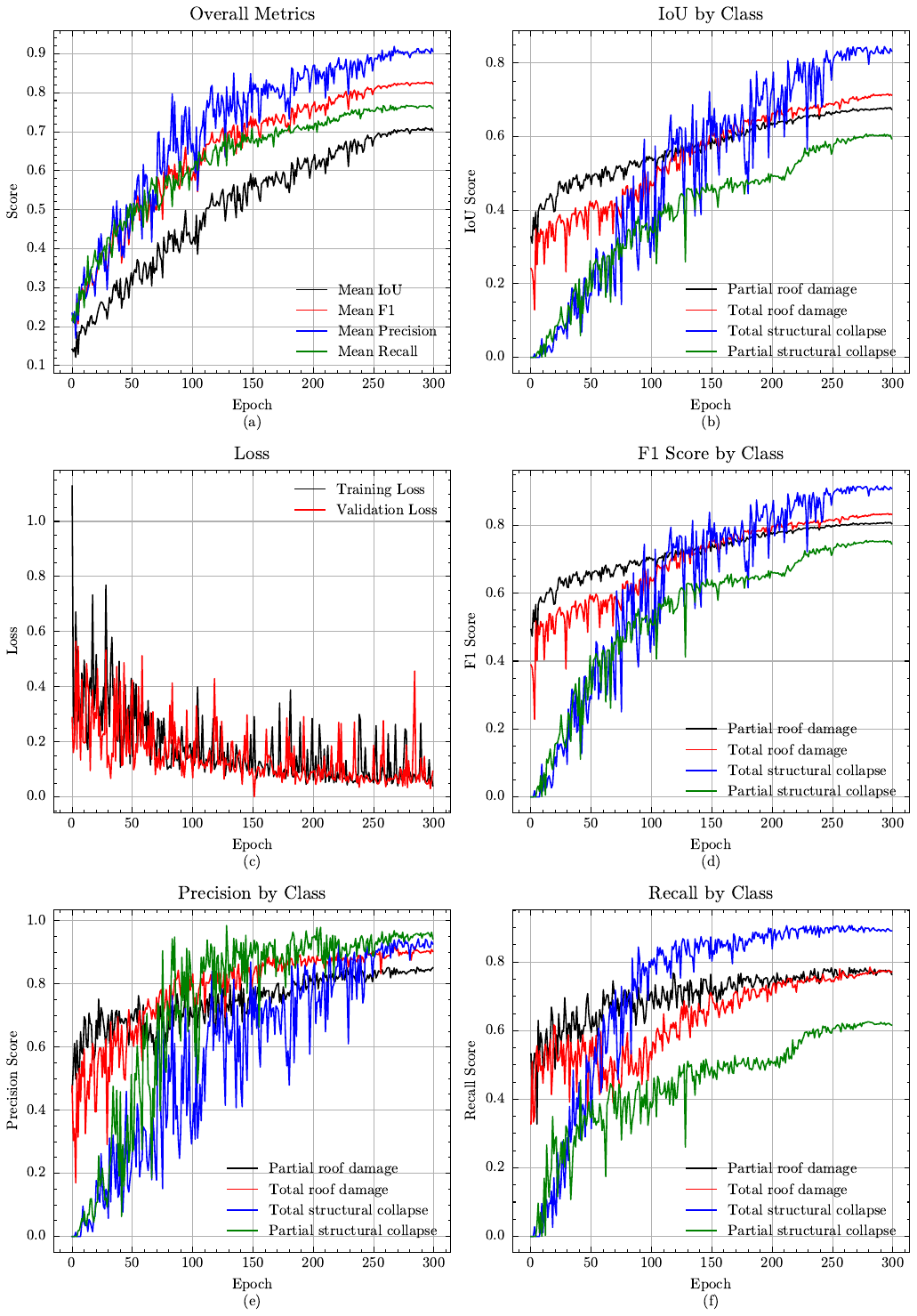}
    \caption{Progression of validation metrics throughout training epochs for Run 2. The plot displays: overall performance (IoU, F1, Precision, Recall), training vs. validation loss, and per-class IoU, F1-score, Precision, and Recall.}\label{fig:app_val_log_metrics_run2}
    
\end{figure}

\begin{figure}[htbp]
    \centering
    \includegraphics[width=\textwidth]{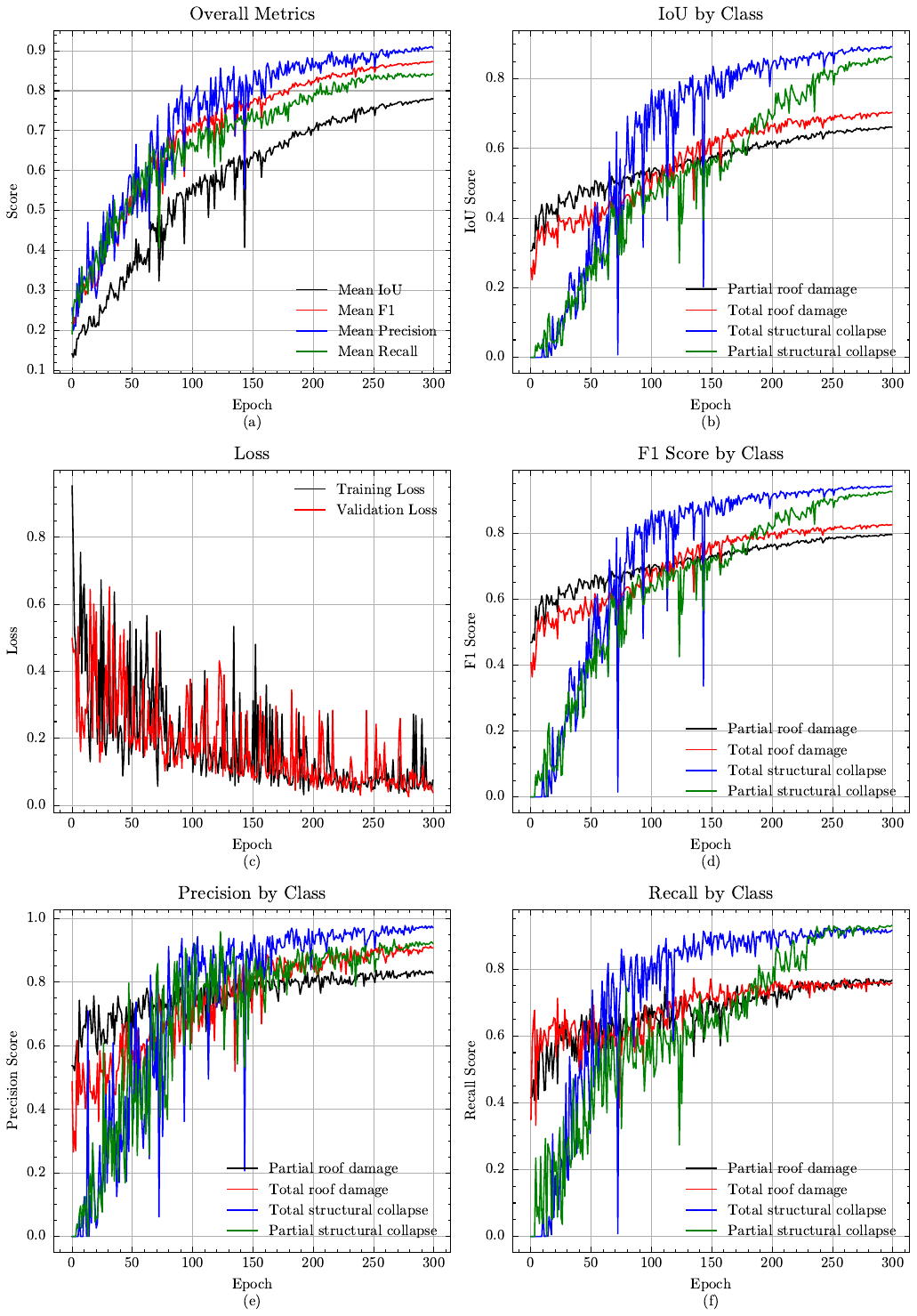}
    \caption{Progression of validation metrics throughout training epochs for Run 3. The plot displays: overall performance (IoU, F1, Precision, Recall), training vs. validation loss, and per-class IoU, F1-score, Precision, and Recall.}\label{fig:app_val_log_metrics_run3}
    
\end{figure}

\begin{figure}[htbp]
    \centering
    \includegraphics[width=\textwidth]{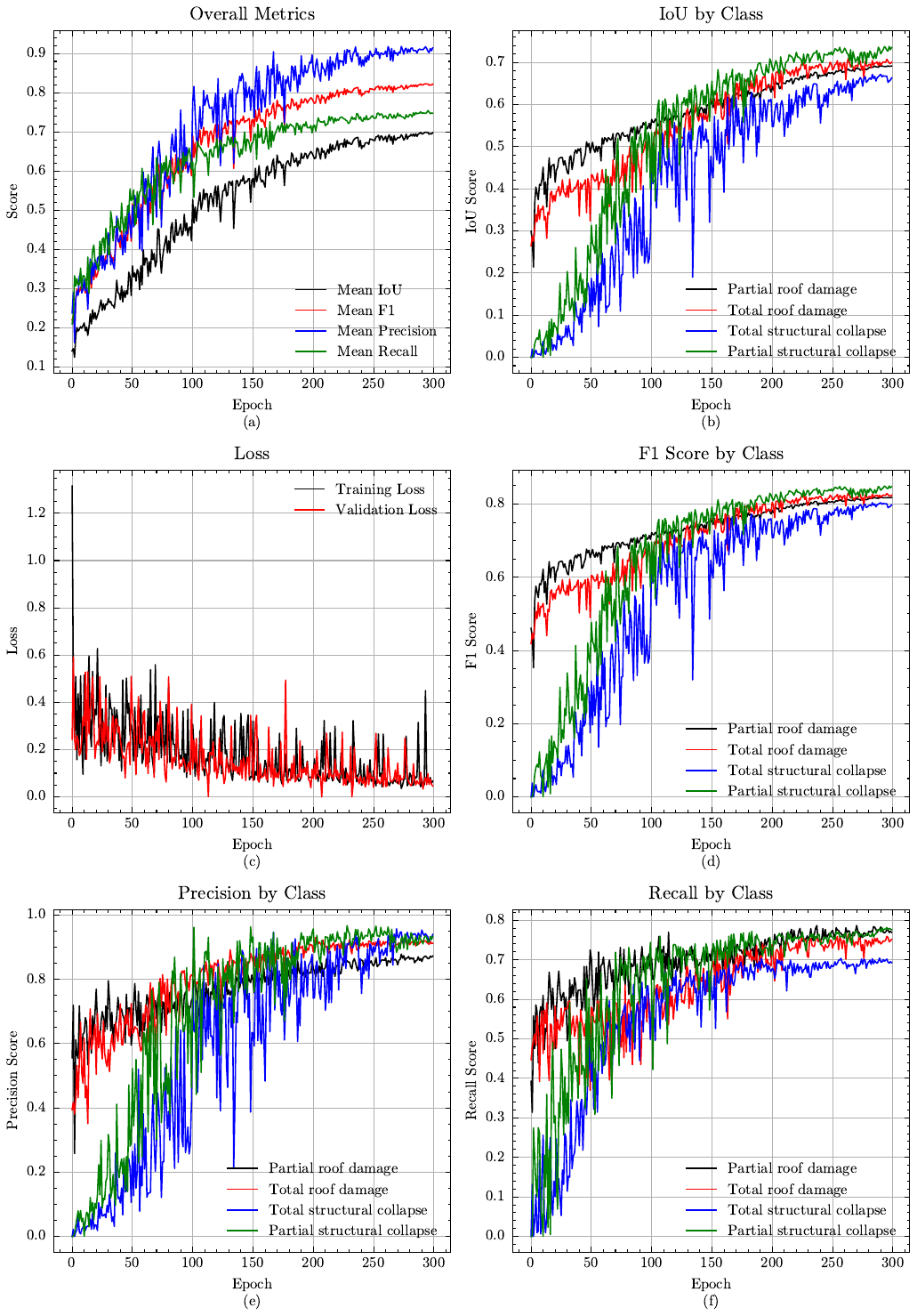}
    \caption{Progression of validation metrics throughout training epochs for Run 4. The plot displays: overall performance (IoU, F1, Precision, Recall), training vs. validation loss, and per-class IoU, F1-score, Precision, and Recall.}\label{fig:app_val_log_metrics_run4}
    
\end{figure}



\clearpage
\begingroup
  \sloppy             
  \bibliography{references}
\endgroup

\end{document}